\documentclass{esannV2}
\usepackage{graphicx}
\usepackage[latin1]{inputenc}
\usepackage{amssymb,amsmath,array}
\usepackage[numbers,sort&compress]{natbib}

\usepackage{bm}
\usepackage{subcaption}

\newcommand{\ReLU}{\text{ReLU}}

\begin{document}
%style file for ESANN manuscripts
\title{On-line learning dynamics of ReLU neural networks using statistical physics techniques}

%***********************************************************************
% AUTHORS INFORMATION AREA
%***********************************************************************
\author{Michiel Straat$^1$ and Michael Biehl$^1$
%
% Optional short acknowledgment: remove next line if non-needed
\thanks{We acknowledge financial support through the Northern Netherlands Region of Smart Factories
(RoSF) consortium,  see http://www.rosf.nl.}
%
% DO NOT MODIFY THE FOLLOWING '\vspace' ARGUMENT
\vspace{.3cm}\\
%
% Addresses and institutions (remove "1- " in case of a single institution)
1- Bernoulli Institute for Mathematics, Computer Science\\ and Artificial Intelligence, University of Groningen \\
Nijenborgh 9, 9747AG Groningen, The Netherlands
%
% Remove the next three lines in case of a single institution
}
%***********************************************************************
% END OF AUTHORS INFORMATION AREA
%***********************************************************************

\maketitle

\begin{abstract}
We introduce exact macroscopic on-line learning dynamics of two-layer neural networks with ReLU units in the form of a system of differential equations, using techniques borrowed from statistical physics. For the first experiments, numerical solutions  reveal similar behavior compared to sigmoidal activation researched in earlier work. In these experiments the theoretical results show good correspondence with simulations. In overrealizable and unrealizable learning scenarios, the learning behavior of ReLU networks shows distinctive characteristics compared to sigmoidal networks.
\end{abstract}

\section{Introduction}
Statistical physics techniques have been used successfully in the theoretical analysis of various machine learning models, including neural networks \cite{Biehl1995, Saad1995, DriftLVQSVM2018} and prototype-based models \cite{Biehl2007,DriftLVQSVM2018}. In the context of neural networks, several learning scenarios have been studied, e.g., on-line gradient descent learning \cite{Biehl1995,Saad95_2,Vicente97,Inoue02}, learning in non-stationary environments \cite{DriftLVQSVM2018} and batch learning \cite{Biehl1998}. Macroscopic quantities, the so-called order parameters of the system, aggregate and summarize the usually large number of individual parameters of the machine learning model. In model situations, Central Limit Theorems (CLT) in combination with the consideration of the thermodynamic limit facilitate an exact description of the macroscopic dynamics in the form of a system of ordinary differential equations (ODE). It provides a useful tool to study the behavior of learning theoretically, in order to gain a deeper understanding of the learning process, which could potentially be used to improve algorithms used in practical scenarios. In the context of deep learning, Rectified Linear Unit (ReLU) activation has become popular mainly due to improved empirical performance compared to sigmoidal activation, e.g., see \cite{Glorot2011}. Here we formulate and study exact macroscopic gradient descent learning dynamics for the Soft Committee Machine (SCM), with the aim of increasing theoretical understanding of the behavior of ReLU activation in neural networks.

\section{Macroscopic ReLU learning dynamics of the SCM}
We consider regression where for an input $\bm{\xi}\in \mathbb{R}^N$ a teacher SCM with $M$ hidden units computes the target output $\tau(\bm{\xi}) \in \mathbb{R}$ and a student SCM with $K$ hidden units computes the hypothesis $\sigma(\bm{\xi}) \in \mathbb{R}$ :
\begin{equation}
	\tau(\bm{\xi}) = \sum_{n=1}^M g(y_n) \quad y_n=\bm{B}_n \cdot \bm{\xi}, \quad \sigma(\bm{\xi}) = \sum_{i=1}^K g(x_i) \quad x_i=\bm{J}_i \cdot \bm{\xi}\, .
\end{equation}
Above, $\bm{B}_n \in \mathbb{R}^N$ and $y_n \in \mathbb{R}$ denote teacher weight vectors and pre-activations, respectively. In case of the student, those are denoted by $\bm{J}_i \in \mathbb{R}^N$ and $x_i \in \mathbb{R}$. We consider for the activation function $g(x)$: $\ReLU(x)=x \theta(x)$, where $\theta(x)$ is the unit step function. The student weights $\bm{J}$ are adaptable and we assume that the teacher weights $\bm{B}$ stay constant, i.e., the target rule remains fixed. In the on-line learning scenario at step $\mu$, a new independent example $\bm{\xi}^\mu$ is presented from a stream. The direct error for $\bm{\xi}^\mu$ and the generalization error are defined as:
\begin{equation}
\epsilon(\bm{J},\bm{\xi}^\mu) = \frac12 (\sigma^\mu - \tau^\mu)^2, \quad \epsilon_g(\bm{J}) = \langle \epsilon(\bm{J},\bm{\xi}) \rangle_{\bm{\xi}}\, ,
\end{equation}
where $\langle \cdot \rangle_{\bm{\xi}}$ denotes averaging over the input distribution.
One estimates the input distribution in practice, but here we consider i.i.d. Gaussian random components $\xi_i \sim \mathcal{N}(0,1)$.

For each presentation $\bm{\xi}^\mu$, the adaptation of the student weight vector $\bm{J}_i$ is guided by gradient descent on $\epsilon(\bm{J}^\mu,\bm{\xi}^\mu)$ with respect to $\bm{J}_i$, resulting in the update rule:
\begin{equation}\label{eq:weightUpdate}
	\bm{J}_i^{\mu+1} = \bm{J}_i^\mu + \frac{\eta}{N} 
	\delta_i^\mu \bm{\xi}^\mu, \quad \delta_i^\mu = (\tau^\mu - \sigma^\mu) g'(x_i^\mu)
\end{equation}
where $\eta$ is the so-called learning rate which is scaled with the input dimension $N$. Note that from Equation \eqref{eq:weightUpdate}, $g(x)$ should be differentiable. $\ReLU'(0)$ is undefined, but in practice one chooses a value for this rare case.

The choice of i.i.d. components $\xi_i$ makes the CLT apply for large input dimension $N$. Hence, for large $N$, the pre-activations $x_i$ and $y_n$ become zero-mean Gaussian variables with properties:
\begin{align}\label{eq:moments_innerfields}
\begin{split}
&\langle x_i x_j \rangle = \bm{J}_i \cdot \bm{J}_j = Q_{ij}, \enspace \langle x_i y_n \rangle = \bm{J}_i \cdot \bm{B}_n = R_{in}, \enspace \langle y_n y_m \rangle = \bm{B}_n \cdot \bm{B}_m = T_{nm}\, .
\end{split}
\end{align}
The variables $R_{in}$, $Q_{ik}$ and $T_{nm}$ are macroscopic variables of the system, so-called \textit{order parameters}. Here we fix the rule properties to $T_{nm}=\delta_{nm}$. Combining the above equations with gradient update Equation \eqref{eq:weightUpdate} yields stochastic update equations for the order parameters directly. In the thermodynamic limit $N\to \infty$, the normalized time variable $\alpha= \mu/N$ can be considered continuous and the order parameters self-average as proved in \cite{Reents98}. Hence, averaging leads to a system of ODEs, e.g., shown in \cite{Saad1995}, describing exact macroscopic dynamics in the thermodynamic limit. For $g(x)=\ReLU(x)$, the system is:
\begin{align}\label{eq:macroscopicReLUSystem}
\begin{split}
\frac{dR_{in}}{d\alpha} &= \eta\left[\sum_{m=1}^M \langle\theta(x_i) y_n y_m \theta(y_m)\rangle - \sum_{j=1}^K \langle\theta(x_i) y_n  x_j \theta(x_j)\rangle\right]\, , \\
\frac{dQ_{ik}}{d\alpha} &= \eta \left[\sum_{m=1}^M \langle \theta(x_i) x_k y_m \theta(y_m) \rangle - \sum_{j=1}^K \langle \theta(x_i) x_k x_j \theta(x_j) \rangle\right] \\
&+ \eta \langle x_k \delta_i \rangle + \eta^2 \langle \delta_i \delta_k \rangle\, , 
\end{split}
\end{align}
where the term $\eta \langle x_k \delta_i \rangle$ in the second equation is the same as the first term for $i$ and $k$ interchanged. The averages of the form $\langle \theta(u) v w \theta(w) \rangle$ are taken with respect to the 3D joint Gaussian distribution $P(\bm{x},\bm{\Sigma})$, for variable vector $\bm{x}=(u,v,w)^T$ and covariance matrix $\bm{\Sigma}=\langle \bm{x} \bm{x}^T \rangle$, which is populated with relevant variances and covariances from Equations \eqref{eq:moments_innerfields}. Integration yields the closed form expression:
\begin{equation}\label{eq:I3}
\langle \theta(u) v w \theta(w) \rangle_{\bm{\xi}} = \frac{\sigma_{12} \sqrt{\sigma_{11} \sigma_{33}-\sigma_{13}^2}}{2\pi \sigma_{11}} + \frac{\sigma_{23} \sin ^{-1}\left(\frac{\sigma_{13}}{\sqrt{\sigma_{11} \sigma_{33}}}\right)}{2\pi} + \frac{\sigma_{23}}{4}\, ,
\end{equation}
where $\sigma_{ij}$ denotes the corresponding element of matrix $\bm{\Sigma}$.
For general $K$ and $M$, the term $\eta^2 \langle \delta_i \delta_k \rangle$ consists of averages of the form $\allowbreak \langle w z \theta(u) \theta(v) \theta(w) \theta(z)\rangle_{\bm{\xi}}$. For now, we only include the $\eta^2$ term for $K=M=1$. For general $K$ and $M$, we study the dynamics for $\eta \to 0$, neglecting the $\eta^2$ term. Combining Equation \eqref{eq:I3} and \eqref{eq:macroscopicReLUSystem} gives the closed form macroscopics of the ReLU SCM.

\section{Experiments}
In this section, we show and discuss for different settings the macroscopic on-line ReLU dynamics as obtained from the theoretical ODEs from Equation \eqref{eq:macroscopicReLUSystem}. Theoretical results are compared with simulations for sufficiently large $N$.

\begin{figure}[h!]
    \centering
    \begin{subfigure}[t]{0.5\textwidth}
        \centering
        \includegraphics[width=\textwidth]{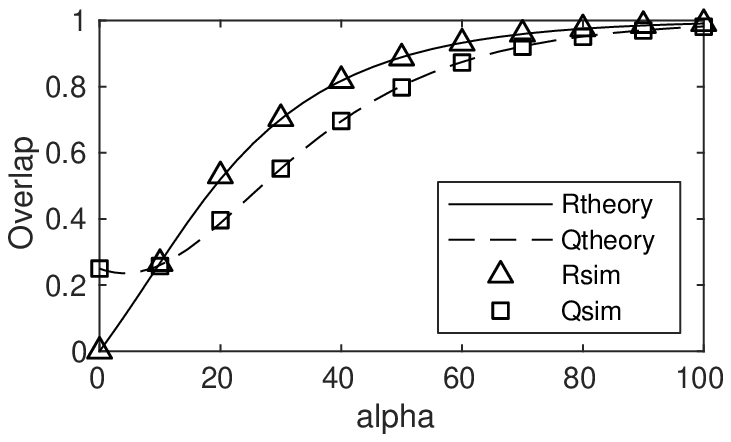}
%        \caption{}
%        \label{fig:perceptron_RQeta01}
    \end{subfigure}%
    ~
    \begin{subfigure}[t]{0.5\textwidth}
        \centering
        \includegraphics[width=\textwidth]{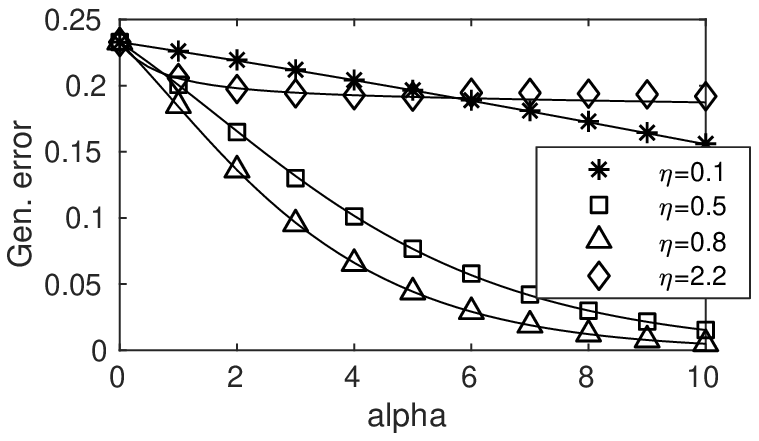}
%        \caption{}
%        \label{fig:perceptron_learningRates}
    \end{subfigure}
    \caption{\textit{Left}: Evolution of order parameters. $R$ and $Q$ with $\eta=0.1$, $R(0)=0$ and $Q(0)=0.25$. \textit{Right}: Evolution of $\epsilon_g$ for different $\eta$. Note the scale of $\alpha$. Lines and symbols show theoretical and simulation results, respectively. $N=1000$ is used in the simulations.}
    \label{fig:ReLUPerceptron}
\end{figure}

We first consider perceptron learning: $M=K=1$. Initial conditions $(R_0, Q_0)=(0,0.25)$ correspond to a random initialization of the student weights $\bm{J}$. For a learning rate $\eta=0.1$, a numerical solution to the ODE system is shown in Figure \ref{fig:ReLUPerceptron}.

One observes an increase in both $R$ and $Q$, indicating increasing similarity of the student to the rule and increasing weight magnitude. The state $(R, Q)=(1,1)$ is the perfect solution that corresponds to equality of student and teacher, i.e., $\bm{J} = \bm{B}$. In fact, $(R, Q)=(1,1)$ is a fixed point of the system for all meaningful $\eta$. Defining $(r,q)=(R-1,Q-1)$, a linearization of the dynamics is $(r',q')^T = \bm{A}(\eta)(r,q)^T$, where $\bm{A}(\eta)$ is the Jacobian in the fixed point. The eigenvalues of $\bm{A}(\eta)$ are given by $\bm{\lambda}=\{-\eta/2, 1/2 \eta^2 - \eta\}$ with corresponding eigenvectors $\bm{u}_1=(1/2, 1)^T$ and $\bm{u}_2=(0, 1)^T$. As $\lambda_2 \geq 0$ for $\eta \geq 2$, it follows that for $\eta < 2$ the fixed point is asymptotically stable and we define this critical learning rate as $\eta_c=2$. Figure \ref{fig:ReLUPerceptron} (right) shows the evolution of $\epsilon_g$ for several $\eta$. Convergence is slow for $\eta<<\eta_c$ but also for $\eta\approx \eta_c$.

\begin{figure}[h!]
	\centering
	\begin{subfigure}[t]{0.5\textwidth}
		\centering
		\includegraphics[width=\textwidth]{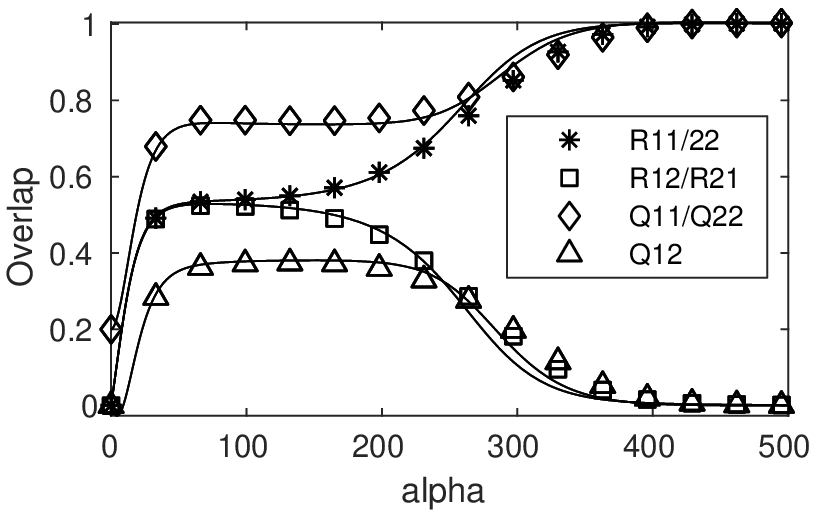}
%		\caption{}
%		\label{fig:ReLUNetworkK2M2Order}
	\end{subfigure}%
	~
	\begin{subfigure}[t]{0.5\textwidth}
		\centering
		\includegraphics[width=\textwidth]{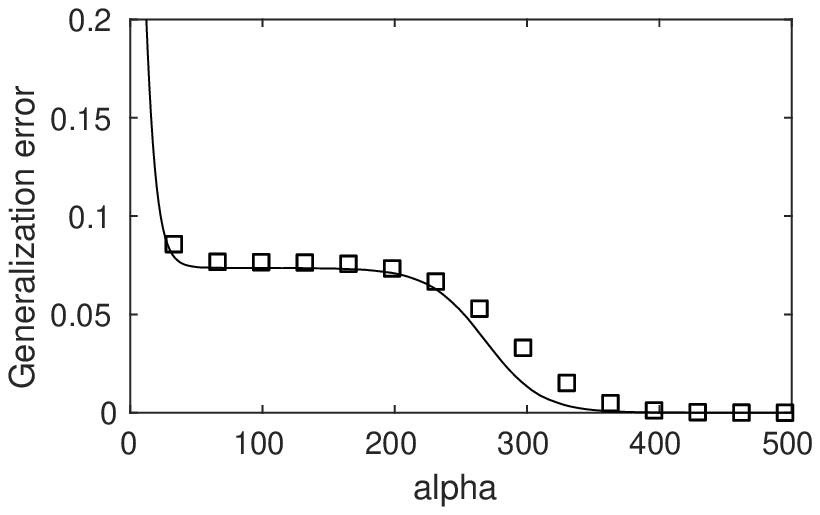}
%		\caption{}
%		\label{fig:ReLUNetworkK2M2GenErr}
	\end{subfigure}
	\caption{\textit{Left}: Evolution of order parameters for the case $K=M=2$ and $\eta=0.1$. \textit{Right}: Evolution of the generalization error. Symbols show simulation results for $N=10^4$.}
	\label{fig:ReLUNetworkK2M2}
\end{figure}

Figure \ref{fig:ReLUNetworkK2M2} shows dynamics for the ReLU network with $K=M=2$. Initial conditions are $R_{in}=10^{-3}\delta_{in}$ and $Q_{11}=0.2$, $Q_{12}=0$, $Q_{22}=0.3$. The learning process is characterized by a suboptimal plateau in which $R_{in} \approx 0.52$ for all $i,n$, i.e., there is no specialization of students towards specific teachers. The symmetric plateaus are a property of learning in soft committee machines\cite{Saad1995,Biehl1995} and they arise due to a repulsive fixed point of the system. An expression for the length of the plateau can be found in \cite{Biehl1996}. From the linearization of the ReLU dynamics, there is one positive eigenvalue that guides the escape: $\lambda_5 = 0.24$ with corresponding eigenvector \textbf{u}\textsubscript{5}=(0.5,-0.5,-0.5,0.5,0,0,0)\textsuperscript{T}: It causes the observed specialization of each student towards one teacher. The onset of specialization is associated with a decrease in generalization error, see Figure \ref{fig:ReLUNetworkK2M2} (right).

\begin{figure}[h!]
	\centering
	\begin{subfigure}[t]{0.5\textwidth}
		\centering
		\includegraphics[width=\textwidth]{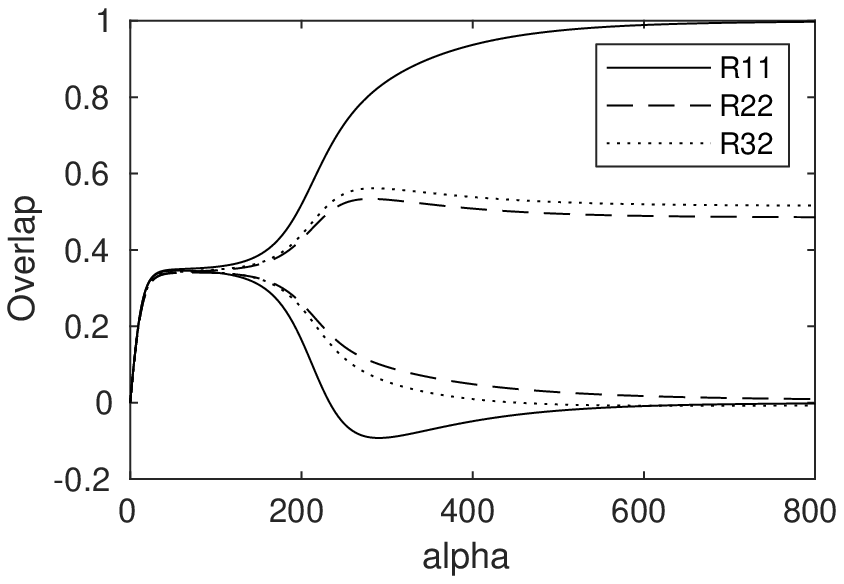}
%		\caption{}
%		\label{fig:ReLUNetworkK2M2Order}
	\end{subfigure}%
	~
	\begin{subfigure}[t]{0.5\textwidth}
		\centering
		\includegraphics[width=\textwidth]{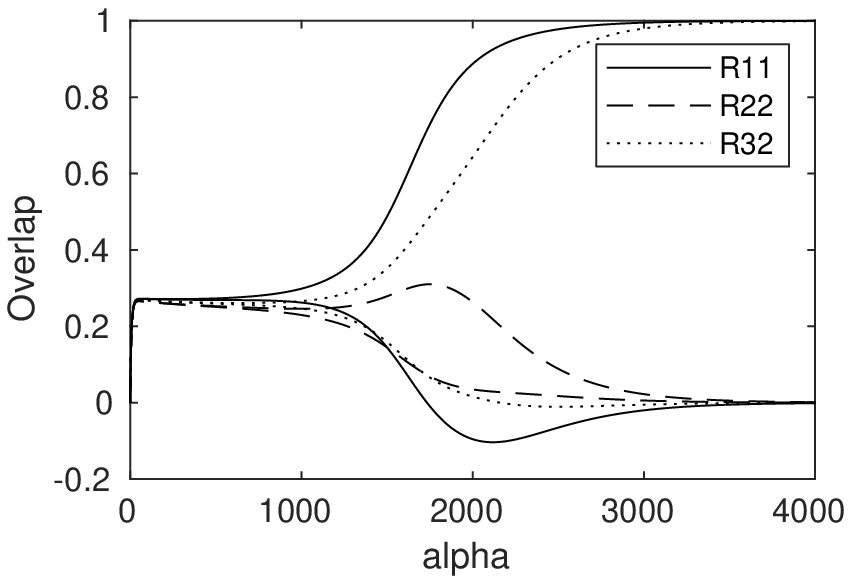}
%		\caption{}
%		\label{fig:ReLUNetworkK2M2GenErr}
	\end{subfigure}
	\caption{Evolution of student-teacher overlap parameters for the case $K=3$ and $M=2$. \textit{Left}: ReLU activation. \textit{Right}: Erf activation. A pair of the same type of curves shows the correlation of one student unit to each of the two teacher units. The legends point to the upper curve of the pair.}
	\label{fig:ReLUNetworkK3M2}
\end{figure}

\begin{table}[h!]
  \centering
  \begin{tabular}{|c|c|c|c|c|c|c|}
    \hline
    & $Q_{11}(\infty)$ & $Q_{12}(\infty)$ & $Q_{13}(\infty)$ & $Q_{22}(\infty)$ & $Q_{23}(\infty)$ & $Q_{33}(\infty)$ \\
    \hline
    ReLU & 1.00 & 0.00 & 0.00 & 0.24 & 0.25 & 0.27 \\
    Erf & 1.00 & 0.00 & 0.00 & 0.00 & 0.00 & 1.00  \\
    \hline
  \end{tabular}
\end{table}

In Figure \ref{fig:ReLUNetworkK3M2} and the table above, results for $K=3$ and $M=2$ are given for ReLU activation (left) and sigmoidal Erf activation (right). For the latter, closed form equations can be found in \cite{Saad1995,Biehl1995}. Non-zero initial conditions are $R_{11}=10^{-3}, Q_{11}=0.2, Q_{22}=0.3,Q_{33}=0.25$. In both cases, $\bm{J}_1$ specializes to $\bm{B}_1$. In the ReLU case, $\bm{J}_2$ and $\bm{J}_3$  achieve a similar overlap with $\bm{B}_2$. From $Q_{22}\approx Q_{33} \approx Q_{23}\approx 0.25$ and $R_{22}\approx R_{32}\approx 0.5$, it follows that $\bm{J}_2=\bm{J}_3 \parallel \bm{B}_2$ i.e., $\bm{J}_2 \approx a\bm{B}_2$ and $\bm{J}_3 \approx b\bm{B}_2$ for $a=b=0.5$ and therefore $\bm{J}_2 + \bm{J}_3 = \bm{B}_2$. Hence, two units of the ReLU student learn both the same teacher unit apart from a scaling and there are in fact infinitely many solutions possible for different $a$ and $b$, $a+b=1$. The observed behavior is a consequence of the piece-wise linear property of the ReLU. Such combinations are not possible for the non-linear Erf: In this case, $R_{22}$ decreases to zero due to $Q_{22}(\alpha \to \infty)=0$, equivalent to $\bm{J}_2=\bm{0}$, effectively removing the unit. In both cases, $\epsilon_g(\alpha \to \infty)=0$ is achieved, since the rule is learned perfectly.

\begin{figure}[h!]
	\centering
	\begin{subfigure}[t]{0.5\textwidth}
		\centering
		\includegraphics[width=\textwidth]{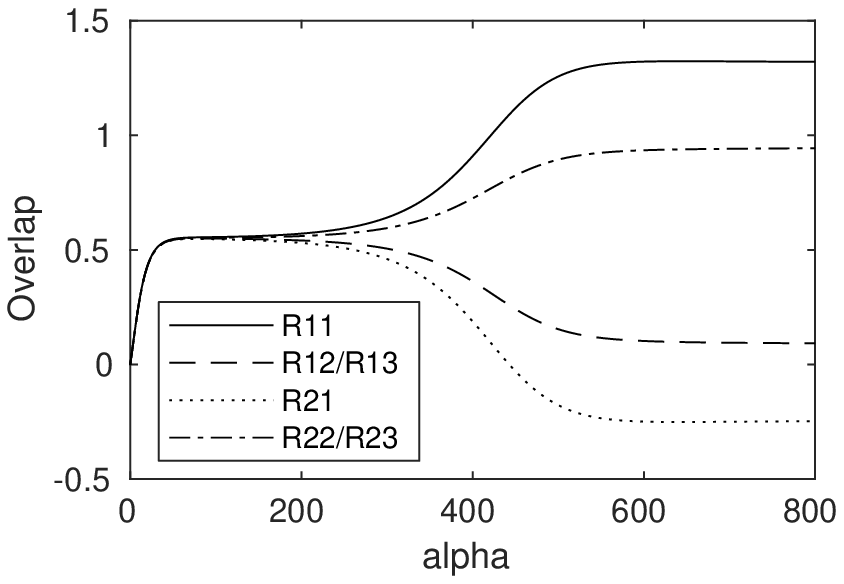}
%		\caption{}
%		\label{fig:ReLUNetworkK2M2Order}
	\end{subfigure}%
	~
	\begin{subfigure}[t]{0.5\textwidth}
		\centering
		\includegraphics[width=\textwidth]{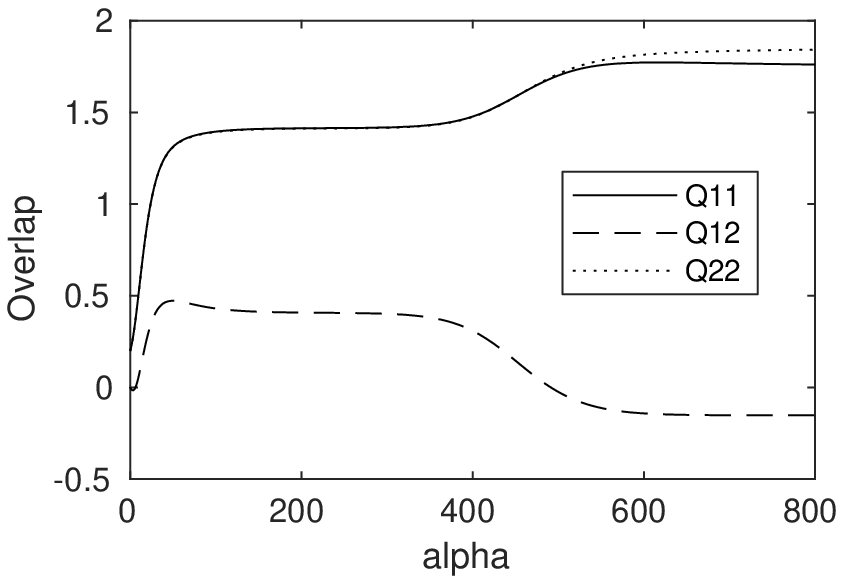}
%		\caption{}
%		\label{fig:ReLUNetworkK2M2GenErr}
	\end{subfigure}
	\caption{Overlaps for the ReLU network with $K=2$ and $M=3$. \textit{Left}: Evolution of student-teacher overlaps. \textit{Right}: Evolution of student-student overlaps.}
	\label{fig:ReLUNetworkK2M3}
\end{figure}

In Figure \ref{fig:ReLUNetworkK2M3}, results of the ReLU network for $K=2$ and $M=3$ are shown. Initial conditions are $R_{11}=10^{-3}$, $R_{in}=0$ for $i,j\not = 1$, $Q_{ii}=0.2$ and $Q_{i\not=j}=0$. $\bm{J}_1$ mainly specializes to $\bm{B}_1$. As $R_{22}=R_{23}=0.94$, it is mainly the case that $\bm{J}_2 \approx a\bm{B}_2 + b\bm{B}_3$ for $a\approx b$. Since the student does not realize the rule, $\epsilon_g(\alpha \to \infty) > 0$.

\section{Discussion}
We have formulated macroscopic learning dynamics of two-layer neural networks for ReLU activation. Simulation results for the perceptron and the network with two hidden units show good correspondence. For the perceptron, the optimal solution corresponds to a fixed point of the equations which becomes unstable at a critical learning rate. Sub-optimal plateaus appear in the networks that correspond to fixed points, of which the repulsion causes eventually specialization. For the overrealizable case, ReLU units are combined to deal with the extra complexity. The $\eta^2$ term that we omitted here should be included in future research to get exact equations for general $\eta$. This would also make possible the study of learning rate adaptation schemes within the framework.

% ****************************************************************************
% BIBLIOGRAPHY AREA
% ****************************************************************************

\begin{footnotesize}

% ----------------------------------------------------------------------------

% IF YOU USE BIBTEX,
% - DELETE THE TEXT BETWEEN THE TWO ABOVE DASHED LINES
% - UNCOMMENT THE NEXT TWO LINES AND REPLACE 'Name_Of_Your_BibFile'

\bibliographystyle{unsrt}
{\footnotesize
\bibliography{biblio}}

\end{footnotesize}

% ****************************************************************************
% END OF BIBLIOGRAPHY AREA
% ****************************************************************************

\end{document}